\documentclass[10pt,twocolumn,letterpaper]{article}

\usepackage{wacv}      % To produce the REVIEW version for the algorithms track
\usepackage[utf8]{inputenc}
\usepackage{amsmath}
\usepackage{nicefrac}
\usepackage{amsfonts}
\usepackage{amssymb}
\usepackage{graphicx}
\usepackage{booktabs}
\usepackage[export]{adjustbox}
\usepackage[table]{xcolor}
\definecolor{oursblue}{RGB}{222,235,247}
\usepackage{geometry}

\usepackage[moderate]{savetrees}
\definecolor{wacvblue}{rgb}{0.21,0.49,0.74}
\usepackage[pagebackref,breaklinks,colorlinks,allcolors=wacvblue]{hyperref}

\def\wacvPaperID{774} % *** Enter the WACV Paper ID here
\def\confName{WACV}
\def\confYear{2027}

\title{CXR-Retrieve: Compositional Text-to-Image Retrieval in Chest Radiography}%: Benchmarking Precision and Negation Awareness}

\author{Tomer Erez$^{1}$
\and Moshe Kimhi$^{1}$
\and Chaim Baskin$^{2}$ \and Ehud Rivlin$^{1}$\\
$^{1}$Technion -- Israel Institute of Technology \quad $^{2}$Ben-Gurion University of the Negev
}

\begin{document}

\maketitle

\begin{abstract}
Large chest radiography archives are difficult to search because most studies are paired only with free-text reports rather than structured clinical annotations. Vision-language models offer a natural interface for text-to-image retrieval, but current biomedical models are primarily optimized for report-to-image matching rather than for satisfying short clinical search queries. This creates an objective mismatch: a model may retrieve images related to words in the query while failing to satisfy the full clinical constraint, especially for conjunctions and negations such as ``atelectasis and no pneumonia.''

We introduce \textsc{CXR-Retrieve}, a structured benchmark for compositional chest X-ray text-to-image retrieval. The benchmark contains 5,159 test images from the official test-split of MIMIC-CXR-JPG~\cite{johnson2019mimicjpg} and 145 textual queries spanning single and conjunction findings, both positive and negative. Relevance is defined by whether a retrieved image satisfies all asserted pathology constraints, rather than by whether it matches a paired report.

We further propose a label-aware contrastive fine-tuning objective for clinical retrieval. Our method attracts image-text pairs with compatible asserted pathology constraints, including shared confirmed absences, while explicitly repelling contradictory pairs. Starting from the in-domain CXR-CLIP checkpoint, our method improves Precision@5 over CXR-CLIP by 8.5 percentage points on two-pathology conjunctions and by 22.0 percentage points on negation queries. These results show that reliable chest X-ray retrieval requires training objectives that model not only which findings are mentioned, but also how they are clinically asserted.
\end{abstract}

\begin{figure}[t]
  \centering
  \includegraphics[width=\linewidth]{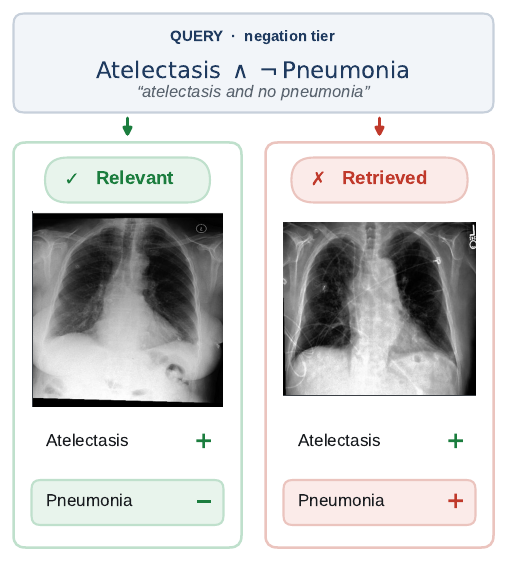}
    \caption{Example where negation query ''image with Atelectasis and no Pneumonia''. CLIP retrieves an image with both pathologies from the database.}
  \label{fig:negation failure}
\end{figure}

\section{Introduction}
Clinical medical archives contain massive volumes of imaging data \cite{Kumar2013, QAYYUM20178, Cheng2024}, yet the vast majority remain unannotated, severely limiting searchable access\cite{Sotomayor2021, Murphy2015-zc} or learning \cite{kimhi2024semi}. Effective zero-shot image retrieval holds profound clinical and research value\cite{Litjens_2017}: it enables rapid dataset construction, supports case-based radiologist education, and powers clinical decision-support tools that surface visually similar prior cases\cite{IGLESIAS2026109540}. While vision-language pretrained models (VLPs) like CLIP~\cite{radford2021learning} have demonstrated remarkable success in learning joint image-text embedding spaces, their application to complex medical domains remains constrained.

A primary bottleneck is a fundamental objective mismatch. Most foundational medical VLPs, such as CXR-CLIP~\cite{you2023cxrclip} and BioMedCLIP~\cite{biomedclip}, are trained almost exclusively under a \textbf{report-matching objective} , aiming to retrieve a study based on its corresponding full radiology report. This is fundamentally distinct from \textbf{clinical query retrieval}, where a user inputs a free-form constraint (e.g., ``pleural effusion but no cardiomegaly'') to retrieve scans satisfying that specific clinical profile. Due to this mismatch, current models exhibit two systematic failure modes. First, \textbf{compositional queries} cause sharp precision degradation\cite{thrush2022winoground, ray2023cola}, where multi-pathology conjunctions overwhelm the model's standard retrieval capacity (Tab \ref{tab:main_results}). Second, \textbf{negation constraints} effectively fail entirely due to a systemic affirmation bias that incorrectly links negated text tokens \cite{alhamoud2025visionlanguagemodelsunderstandnegation} to positive visual features  (Tab \ref{tab:main_results}).

To study this gap, we formulate chest X-ray text-to-image retrieval as
constraint satisfaction over clinical findings. A retrieved image is
relevant only if it satisfies all asserted query constraints, including
both positive findings and explicit exclusions. This setting exposes
failure modes that are largely hidden by report-level retrieval metrics.
\textbf{Our contributions are threefold:}

\textbf{(i)} We introduce \textsc{CXR-Retrieve}, a structured benchmark for
single-finding, conjunctive, and negation-aware chest X-ray retrieval
derived from MIMIC-CXR-JPG labels.

\textbf{(ii)} We propose a conflict-aware contrastive objective that aligns
image and text embeddings according to asserted pathology constraints:
compatible captions provide additional positives, while contradictory
caption-image pairs are explicitly repelled.

\textbf{(iii)} We provide a detailed evaluation against general-domain,
biomedical, and in-domain chest X-ray CLIP baselines, showing that
standard report-trained models degrade sharply on conjunction and
negation queries, whereas our method substantially improves retrieval
precision on these harder query types.

\begin{figure*}
    \centering
    \includegraphics[width=\linewidth]{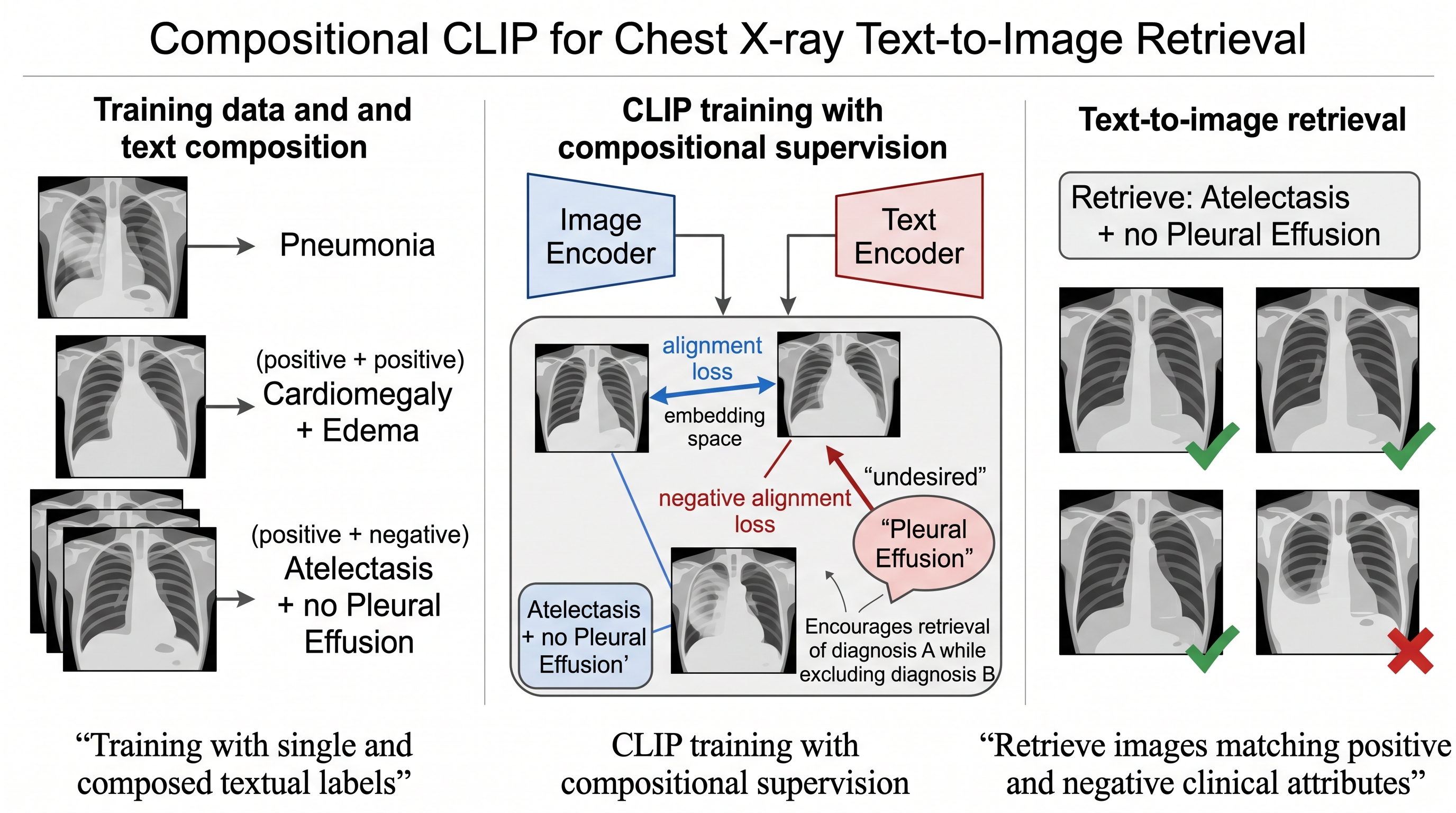}
\caption{\textbf{Overview of the proposed label-aware contrastive framework.} 
Traditional medical vision-language models match dense radiology reports directly to images, failing on short, precise clinical queries. 
In contrast, our method uses structured pathology labels to guide the alignment space. 
Given synthesized clinical text queries (single, conjunctive, or negative), we compute a text-label matrix to evaluate clinical agreement via a custom label kernel. 
Compatible image-text pairs (sharing confirmed findings or exclusions) are brought closer together using soft contrastive targets (\emph{Attraction}). 
Crucially, pairs that present explicit clinical contradictions (e.g., asserting a finding present vs. absent) are actively pushed apart via hinge loss (\emph{Repulsion/Hard Negative Mining}). 
This assertion-driven supervision forces the joint space to respect clinical logic and negation constraints rather than simple token co-occurrence.}
\label{fig:method}
\end{figure*}

\section{Related Work}

\paragraph{Vision-language pretraining for medical imaging.}
CLIP-style pretraining has become a common approach for learning transferable image-text representations from large paired datasets fast and efficiently \cite{radford2021learning,wave}. In medical imaging, models %such as ConVIRT~\cite{zhang2022contrastivelearningmedicalvisual}, GLoRIA~\cite{huang2021gloria}, MedCLIP~\cite{wang2022medclipcontrastivelearningunpaired}, BioMedCLIP~\cite{biomedclip}, and CXR-CLIP~\cite{you2023cxrclip} 
adapt this paradigm to biomedical or radiology data \cite{zhang2022contrastivelearningmedicalvisual,huang2021gloria,wang2022medclipcontrastivelearningunpaired,biomedclip,you2023cxrclip}.
While those models are both efficient and improve in-domain alignment, they are trained and evaluated with image-caption or image-report matching objectives.
Such objectives are not designed to evaluate whether a retrieved image satisfies a short clinical quarry, and more importantly, fail to retrieve correctly multiple findings or explicit exclusions.

\vspace{-0.6em}
\paragraph{Chest X-ray retrieval.}
Chest X-ray imaging is a widely performed and extensively digitized medical imaging modality, supported by several large-scale public datasets~\cite{PhysioNet-mimic-cxr-2.1.0, irvin2019chexpert, Wang_2017}. These datasets enable large-scale retrieval experiments over real clinical archives.
Radiology retrieval methods commonly use paired reports, mined report
labels, or global image-text similarity to retrieve clinically related
studies~\cite{zhang2025radirscalableframeworkmultigrained, you2023cxrclip,endo2021retrieval, HAQ2021101847, jeong2023multimodalimagetextmatchingimproves}.  These methods are useful for report-level matching and
broad semantic search, but they do not directly test whether the retrieved
image satisfies a user-specified pathology query. In contrast,
\textsc{CXR-Retrieve} formulates chest X-ray retrieval as constraint
satisfaction: a retrieved image is relevant only when it satisfies every
positive and negative condition in the query.

\vspace{-0.6em}
\paragraph{Compositionality and negation in vision-language models.}
Vision-language models are known to struggle with compositional
understanding. Benchmarks such as Winoground~\cite{thrush2022winoground}, COLA~\cite{ray2023cola}, and
ARO~\cite{yuksekgonul2023visionlanguagemodelsbehavelike} show that models often recognize individual concepts while
failing to bind them into the correct joint configuration. Negation is
especially challenging because the negated concept is still explicitly
mentioned in the text, causing models to retrieve images containing the
forbidden concept, a failure characterized by the CLIP negation benchmark~\cite{alhamoud2025visionlanguagemodelsunderstandnegation}.
Chest radiography is a clinically important setting for this problem because reports routinely distinguish affirmed, negated, uncertain, and unmentioned findings. Our
work addresses this issue by using explicit pathology assertions to
define compatible and contradictory image-text pairs during training.

\section{Benchmark}

We build \textsc{CXR-Retrieve} from MIMIC-CXR~\cite{PhysioNet-mimic-cxr-2.1.0} - 377{,}110 chest radiographs from 227{,}827 studies, paired only with their free-text reports
and no structured labels - which is why models trained to match these reports
to images fail when queried for specific pathologies on demand. We build on the
MIMIC-CXR-JPG~\cite{johnson2019mimicjpg} variant, which adds 14 pathology labels
per study extracted from the reports by two medical-NLP labelers: NegBio ~\cite{peng2017negbiohighperformancetoolnegation} and CheXpert~\cite{irvin2019chexpert}; We used Chexpert~\cite{irvin2019chexpert} labels because it achieved higher accuracy when validated by a radiologist in MIMIC-CXR-JPG~\cite{johnson2019mimicjpg}. 
This allows us to train models to match label-based queries to clinical images. 
The test set has 5{,}159 images and 145 queries.
We keep 10 of the 14 labels, dropping ''Support Devices''
(not a clinical finding), ''No Finding'', ''Enlarged Cardiomediastinum'', and ''Pleural Other'' for low labeler precision.

\subsection{Text-to-Image Retrieval with CLIP}
We pre-compute all image embeddings offline. At query time, the model encodes the text query and retrieves the top-$K$ images whose embeddings have the highest cosine similarity with the query embedding. A query is successful if the retrieved image's labels satisfy all constraints of the query.

\subsection{Label Semantics}
Each label takes one of four values: positive (1), negative (0), uncertain
($-1$), or unmentioned (blank). 
This lets us construct query--image pairs over a single label or a composite of several labels, each required to be present or absent.

\vspace{-0.6em}
\paragraph{Negation evaluation under missing labels.}
Our main benchmark follows a pragmatic retrieval convention: a negated
constraint ``no \(B\)'' is satisfied whenever \(B\) is not explicitly
confirmed positive. This reflects how radiology reports are commonly used
for retrospective cohort construction, but it may overestimate negation
performance when unmentioned findings are truly unknown. We therefore
also report a strict negation setting in which ``no \(B\)'' requires a
confirmed negative CheXpert label. The strict setting reduces the number
of relevant images but provides a conservative estimate of negation-aware
retrieval performance.

\begin{table}[ht]
\centering
\caption{Distribution of CheXpert labels across all samples and pathologies. The dataset is dominated by missing values.}
\label{tab:overall_label_distribution}
\begin{tabular}{lrr}
\hline
\textbf{Label} & \textbf{Count} & \textbf{Percentage} \\
\hline
Positive ($1$)   & 633,824   & 12.00\% \\
Negative ($0$)   & 249,448   & 4.72\% \\
Uncertain ($-1$) & 112,555   & 2.13\% \\
Missing (NaN)    & 4,283,713 & 81.13\% \\
\hline
Total            & 5,279,540 & 100.00\% \\
\hline
\end{tabular}
\end{table}

\paragraph{Caption Synthesis}\label{par:caption_synth}
We construct structured free-text queries from categorical labels to simulate clinical retrieval conditions. For each case, we generate 145 queries spanning three regimes: 10 single-pathology queries (e.g., ``Edema''), 45 pairwise conjunctions (e.g., ``Edema and consolidation''), and 90 negation queries (e.g., ``Edema and no consolidation'').

\subsection{Evaluation Metrics}
We report three metrics:

\noindent\textbf{Precision@$k$:} fraction of the top-$k$ retrieved images that satisfy all constraints of the query, macro-averaged over all queries:
\begin{equation}
    P@k = \frac{1}{|Q|}\sum_{q\in Q}\frac{1}{k}\sum_{i=1}^{k}\mathbf{1}\!\left[\text{image}_i^q \in \mathcal{R}_q\right],
\end{equation}
where $\mathcal{R}_q$ is the set of relevant images for query-$q$.

\noindent\textbf{Recall@$k$:} following standard text-to-image retrieval practice~\cite{radford2021learning}. binary indicator equal to 1 if at least one relevant image appears in the top-$k$ results:
    \begin{equation}
        R@k = \frac{1}{|Q|}\sum_{q\in Q}\mathbf{1}\!\left[\mathrm{top\text{-}}k(q)\cap\mathcal{R}_q\neq\varnothing\right],
    \end{equation}
    where $\mathcal{R}_q$ is the set of relevant images for query $q$.

\noindent\textbf{Hard Negative Retrieval Rate (HNRR)@$k$:} applied to negation queries only. For a query ``\textit{A} and no \textit{B}'', a \emph{hard negative} is a retrieved image where both pathology $A$ and pathology $B$ are confirmed present (label\,=\,1) - i.e.\ it satisfies the positive constraint while directly violating the negation. HNRR@$k$ is the fraction of the top-$k$ results that are hard negatives, macro-averaged over all negation queries:
\begin{equation}
\mathrm{HNRR}@k = \frac{1}{|Q|} \sum_{q\in Q} \frac{1}{k} \sum_{i=1}^{k} \mathbf{1} \!\left[ \mathrm{image}_i^q \in \mathcal{H}_q \right],
\label{eq:hnrr}
\end{equation}

where $\mathcal{H}_q$ denotes the set of hard negatives for query $q$, i.e., images that satisfy the positive constraint while violating the negated pathology. Lower HNRR@k is better, and a value of zero indicates that no retrieved image violates the negation constraint.

\begin{figure}[t]
  \centering
  \includegraphics[width=\linewidth]{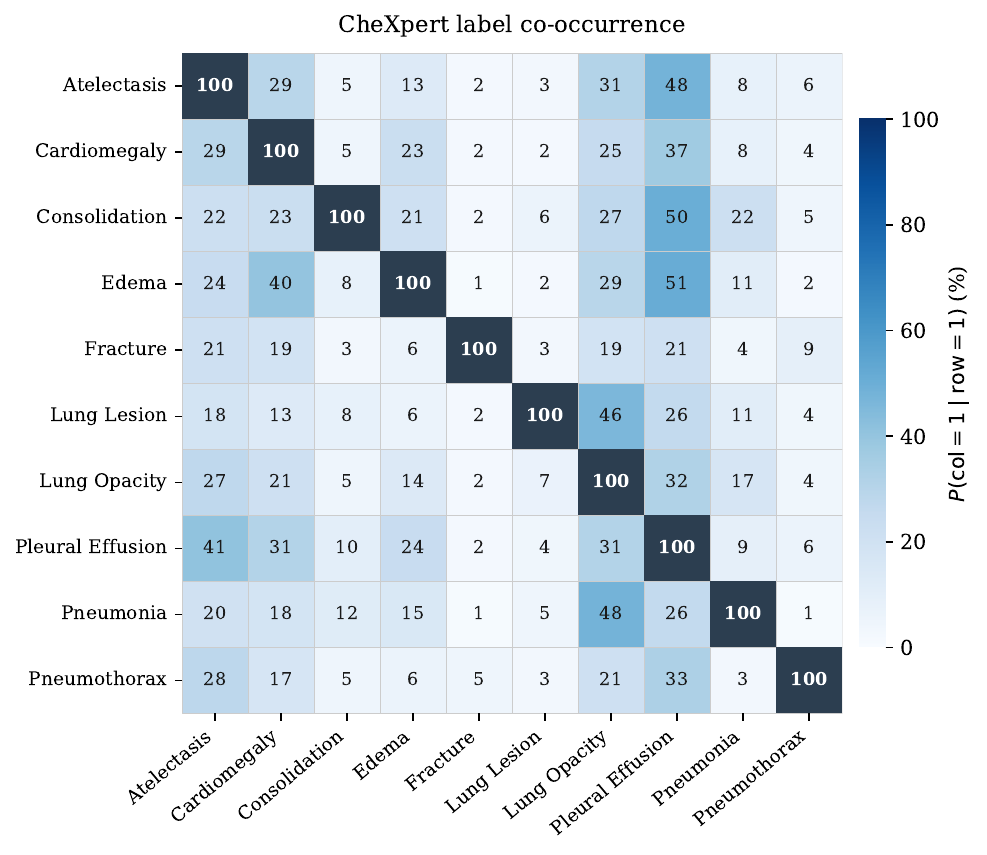}
  \caption{CheXpert label conditional co-occurrence matrix(\%). Cell $(i,j)$ shows the percentage of
samples with label $i$ positive that also have label $j$ positive, i.e.\ $P(\text{col}=1 \mid \text{row}=1)$. Diagonal cells are always 100\,\%.}
  \label{fig:cooccurrence}
\end{figure}

\section{Method}

As outlined in Figure~\ref{fig:method}, we fine-tune a dual-encoder CLIP model to prioritize clinical assertions over token co-occurrence. We achieve this by defining a label-agreement kernel (Sec. \ref{subsec:main_contrast}) that yields soft contrastive targets for attracting compatible pairs, while explicitly repelling clinical contradictions.

\subsection{Label-Aware Contrastive Loss}\label{subsec:main_contrast}

Each training sample $i$ carries a label vector
$\mathbf{y}_i \in \{-1,0,1\}^L$ derived from its synthesized training caption:
labels affirmed in the caption are set to $+1$, the negated label (if present)
to $-1$, and all other entries to $0$.
$K(i,j)$ therefore measures agreement on what each caption asserts rather than
on the complete pathology profiles of the images, directly aligning the training
signal with query-time semantics.
Let $\mathbf{z}_i^v, \mathbf{z}_i^t\in\mathbb{R}^d$ be the $\ell_2$-normalised image
and text embeddings, and $\tau$ the learned logit scale.

\paragraph{Label kernel.}
We measure the semantic similarity of two samples by the dot product of their label
vectors:
\begin{equation}
\begin{split}
K(i,j)
&=
\mathbf{y}_i^\top \mathbf{y}_j
\\
&=
\underbrace{\left|\{l:y_{il}=y_{jl}=1\}\right|}_{\text{shared positives}}
+
\underbrace{\left|\{l:y_{il}=y_{jl}=-1\}\right|}_{\text{shared negatives}}
\\
&\quad-
\underbrace{\left|\{l:y_{il}y_{jl}=-1\}\right|}_{\text{conflicts}}.
\end{split}
\label{eq:kernel}
\end{equation}
$K(i,j)>0$ means the images agree on which findings are present and absent;
$K(i,j)<0$ flags a clinical contradiction.

Shared confirmed absences contribute positively to $K$ by deliberate design:
a radiologist-confirmed absence is a genuine visual feature, and negation query
training requires this signal --- when the caption is ``edema and no consolidation'',
images sharing a confirmed absence of consolidation must attract so the model learns
what ``no~X'' looks like.
Crucially, confirmed absences occupy only $4.7\%$ of label slots (vs.\ $12\%$
positive, $81\%$ unmentioned), so shared absences are rare per batch pair and
cannot dominate $K$.

\paragraph{Soft retrieval targets.}
We zero out conflicting (negative) entries and row-normalize the remaining
non-negative agreement scores to obtain soft contrastive targets:
\[
T_{ij} =
\frac{\max(K_{ij},\,0)}
{\sum_{k=1}^{B} \max(K_{ik},\,0)} .
\]
The self-clamp $K(i,i)\!\leftarrow\!\max(K(i,i),1)$ guarantees that every
image remains paired with its own caption, while semantically compatible
non-identical samples can also serve as positives.

\begin{equation}
\begin{split}
\mathcal{L}
= -\frac{1}{2B}\sum_i \Biggl[
&\sum_j T_{ij}
\log
\frac{
e^{\tau\langle \mathbf{z}_i^v,\mathbf{z}_j^t\rangle}
}{
\sum_k e^{\tau\langle \mathbf{z}_i^v,\mathbf{z}_k^t\rangle}
}
\\
+&
\sum_j T_{ji}
\log
\frac{
e^{\tau\langle \mathbf{z}_i^t,\mathbf{z}_j^v\rangle}
}{
\sum_k e^{\tau\langle \mathbf{z}_i^t,\mathbf{z}_k^v\rangle}
}
\Biggr].
\end{split}
\label{eq:loss}
\end{equation}
Pairs with $T_{ij}>0$ attract in proportion to label agreement. All other pairs -- including clinically conflicting ones ($K(i,j)<0$) -- receive $T_{ij}=0$ and enter the softmax denominator as implicit negatives.

\paragraph{Hard Negative Mining.}
Pairs with $K(i,j)<0$ indicate a direct clinical contradiction: one caption asserts a finding present while the other asserts the same finding absent.
Although these pairs already act as implicit negatives in the softmax denominator of Eq.~\ref{eq:loss}, we additionally apply an explicit repulsion hinge to sharpen their separation.
Let $\mathcal{C} = \{(i,j) : K(i,j) < 0\}$ be the set of conflict pairs.
The hard negative mining loss is
\begin{equation}
\mathcal{L}_{\mathrm{HNM}} =
\frac{1}{|\mathcal{C}|}
\sum_{(i,j)\in\mathcal{C}}
\max\!\left(0,\;\tau\langle\mathbf{z}_i^v,\mathbf{z}_j^t\rangle \right),
\label{eq:hnm}
\end{equation}
The full training objective combines the contrastive attraction loss with this auxiliary repulsion:
\begin{equation}
\mathcal{L}_{\mathrm{total}} = \mathcal{L} + \lambda_{\mathrm{HNM}}\,\mathcal{L}_{\mathrm{HNM}},
\label{eq:total}
\end{equation}
with $\lambda_{\mathrm{HNM}}$ chosen to keep the repulsion weaker than the contrastive attraction in Eq.~\ref{eq:loss}.

\paragraph{Encoding unmentioned labels as unknown.}
Of all label slots in MIMIC-CXR \cite{johnson2019mimicjpg}, $81\%$ are unmentioned (NaN) as seen in table \ref{tab:overall_label_distribution}.
Encoding these as absent ($-1$) would make nearly every batch pair appear to conflict --- an image with confirmed edema would conflict with any image that simply never mentions edema --- flooding $K$ with spurious negative signal.
We therefore encode unmentioned labels as $0$ (unknown) in $\mathbf{y}_i$, so only the $4.7\%$ (table \ref{tab:overall_label_distribution}) of slots carrying a confirmed absence contribute to $K$.
Note the distinction from evaluation: at training time we exclude unknown slots from the supervision signal entirely; at evaluation time, those same missing labels are treated as satisfying a ``no $B$'' constraint, reflecting the clinical assumption that unmentioned findings are likely absent.

  \section{Experiments \& Results}

In this section, we evaluate the proposed \textsc{CXR-Retrieve} framework to answer the following research questions:
\noindent How do report-trained models perform on conjunctive and negation queries relative to single-finding queries, and how does it compare to label-aware attraction and hard-negative mining? (Section \ref{subsec:main})

\noindent Did we learn robust semantic understanding of clinical exclusion, or simply overfit to the linguistic templates synthesized during training? (Section \ref{subsec:negate})

\noindent What is the relative contribution of each proposed system component to the overall retrieval performance? (Section \ref{subsec:ablation})

To investigate these questions, we compare against three baselines: vanilla CLIP~\cite{radford2021learning} (general-domain), BioMedCLIP~\cite{biomedclip} (broad biomedical pretraining), and CXR-CLIP SwinT~\cite{you2023cxrclip} (report-matching on the same MIMIC-CXR data, our primary in-domain baseline). All models have comparable parameter counts ($\sim$136--154M). We report macro-averaged Precision@$K$ and Recall@$K$; since R@1 $\equiv$ P@1 we omit it for brevity.

\subsection{Implementation Details}

Dual-encoder architecture: SwinT image encoder\cite{liu2021swintransformerhierarchicalvision} + Bio\_ClinicalBERT \cite{alsentzer2019publiclyavailableclinicalbert} text encoder, initialized from the CXR-CLIP SwinT~\cite{you2023cxrclip} checkpoint and fine-tuned via LoRA adapters~\cite{hu2021loralowrankadaptationlarge} of rank 12. $\sim$2\% of the model parameters are trainable ($\sim$ 3M of 138M total).
For the hard negative mining weight component in the loss We use $\lambda_{\mathrm{HNM}} = 0.3$. 

\vspace{-7pt}
\paragraph{Training captions.}
During training, captions are sampled per image with the following proportions: 50\% single-pathology captions, 25\% pathology-pair captions, and 25\% negation captions, using the query templates described in Sec.~\ref{par:caption_synth}. Negation captions are generated only for pathologies with a confirmed absent label (CSV value 0).

\begin{figure}[t]
  \centering
  \includegraphics[width=\linewidth]{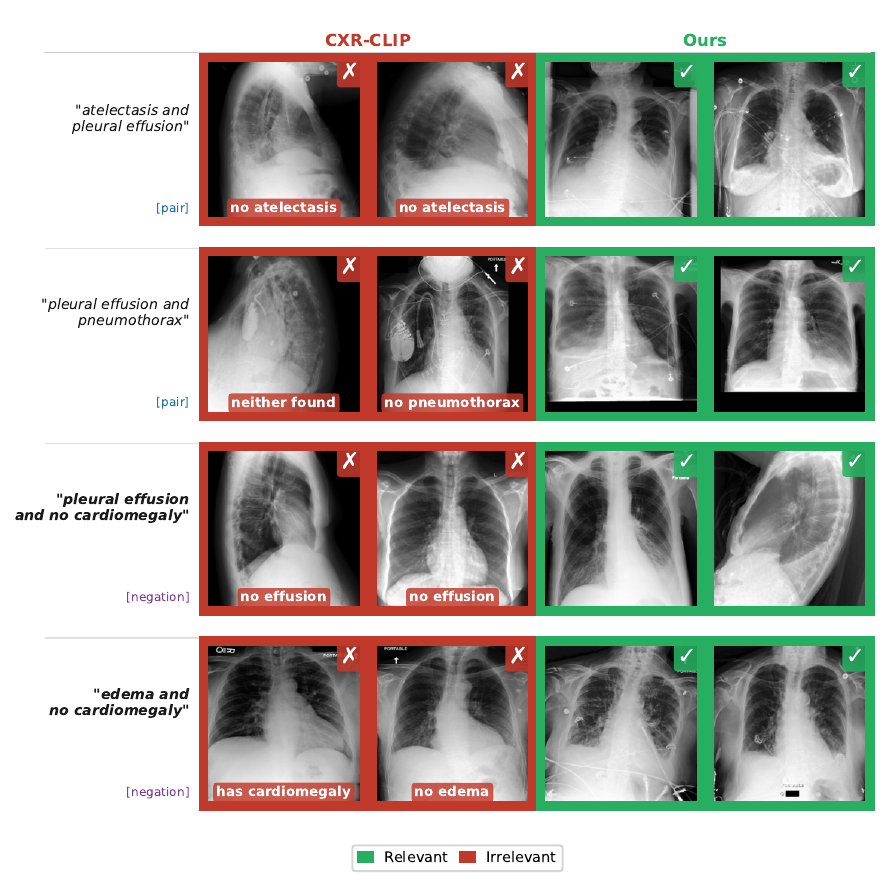}
\caption{Top-2 retrievals on conjunctive (\emph{top}) and negation
  (\emph{bottom}) queries. CXR-CLIP fails on all four. Our model satisfies
  all constraints.}  \label{fig:comparison_collage}
\end{figure}

\begin{figure}[h]
  \centering
  \includegraphics[width=\linewidth]{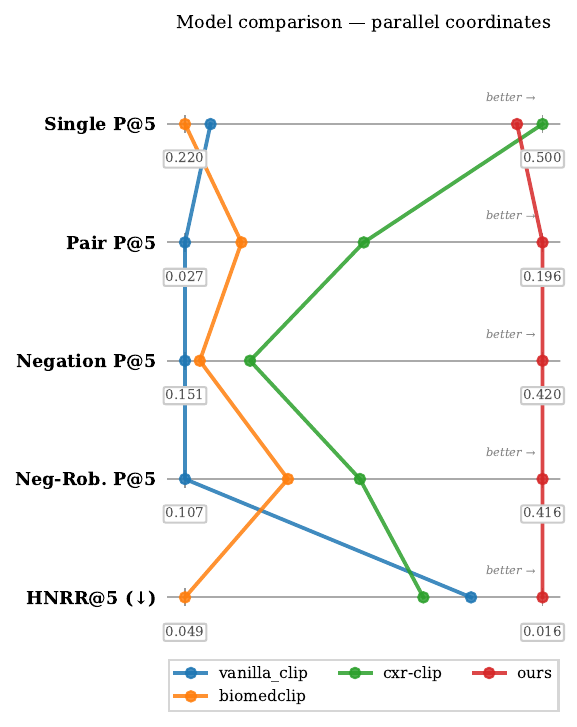}
  \caption{Parallel coordinates comparison across five retrieval metrics. Each axis is normalized to $[0,1]$ (0\,=\,worst, 1\,=\,best); HNRR@5 is inverted since lower is better.}
  \label{fig:parallel_coords}
\end{figure}

%--------------------------------------------------------------
\subsection{Main Retrieval Results}\label{subsec:main}
%--------------------------------------------------------------
 
\begin{table}[t]
\centering
\centering
\caption{Macro-averaged retrieval performance (\%) across the three query
regimes: single-pathology ($N{=}10$), conjunctive pair ($N{=}45$), and negation
($N{=}90$). HNRR@5 (lower is better) is defined only for negation queries. Our
label-aware objective trades a small amount of single-label precision for large
gains on the harder compositional and negation regimes.}
\label{tab:main_results}
\begin{adjustbox}{width=\linewidth}
\begin{tabular}{lcccccc}
\toprule
Model & P@1 & P@3 & P@5 & R@3 & R@5 & HNRR@5\,$\downarrow$ \\
\midrule
\multicolumn{7}{l}{\textit{Single-pathology} (``\textit{A}'')} \\
\cmidrule(lr){1-7}
\texttt{clip}       & 30.0 & 26.7 & 24.0 & 50.0 & 60.0 & -- \\
\texttt{biomedclip} & 20.0 & 20.0 & 22.0 & 40.0 & 50.0 & -- \\
\texttt{cxr-clip}   & 40.0 & 43.3 & \textbf{50.0} & \textbf{90.0} & \textbf{100.0} & -- \\
\rowcolor{oursblue}
\texttt{ours}       & \textbf{50.0} & \textbf{46.7} & 48.0 & \textbf{90.0} & 90.0 & -- \\
\midrule
\multicolumn{7}{l}{\textit{Conjunctive pair} (``\textit{A} and \textit{B}'')} \\
\cmidrule(lr){1-7}
\texttt{clip}       &  6.7 &  2.2 &  2.7 &  6.7 & 11.1 & -- \\
\texttt{biomedclip} &  0.0 &  3.0 &  5.3 &  8.9 & 20.0 & -- \\
\texttt{cxr-clip}   & 11.1 &  9.6 & 11.1 & 22.2 & 33.3 & -- \\
\rowcolor{oursblue}
\texttt{ours}       & \textbf{24.4} & \textbf{20.0} & \textbf{19.6} & \textbf{37.8} & \textbf{51.1} & -- \\
\midrule
\multicolumn{7}{l}{\textit{Negation} (``\textit{A} and no \textit{B}'')} \\
\cmidrule(lr){1-7}
\texttt{clip}       & 14.4 & 15.2 & 15.1 & 34.4 & 45.6 & 2.2 \\
\texttt{biomedclip} & 21.1 & 16.7 & 16.2 & 33.3 & 43.3 & 4.9 \\
\texttt{cxr-clip}   & 21.1 & 20.0 & 20.0 & 43.3 & 54.4 & 2.7 \\
\rowcolor{oursblue}
\texttt{ours}       & \textbf{48.9} & \textbf{47.0} & \textbf{42.0} & \textbf{82.2} & \textbf{90.0} & \textbf{1.6} \\
\bottomrule
\end{tabular}
\end{adjustbox}
\end{table}

We interpret single-pathology P@1 cautiously due to the small number of query types (N=10), and emphasize pairwise and negation regimes as the primary evaluation focus.

On single-pathology retrieval (table \ref{tab:main_results}), our method remains competitive with the strongest in-domain baseline, CXR-CLIP. We achieve the best P@1 (50.0) and P@3 (46.7), while CXR-CLIP retains a slight advantage at P@5 (50.0 vs. 48.0) and R@5 (100.0 vs. 90.0). The proposed objective preserved single-label retrieval performance despite being optimized for more complex queries.

\begin{figure}[h]
\centering
\includegraphics[width=\linewidth]{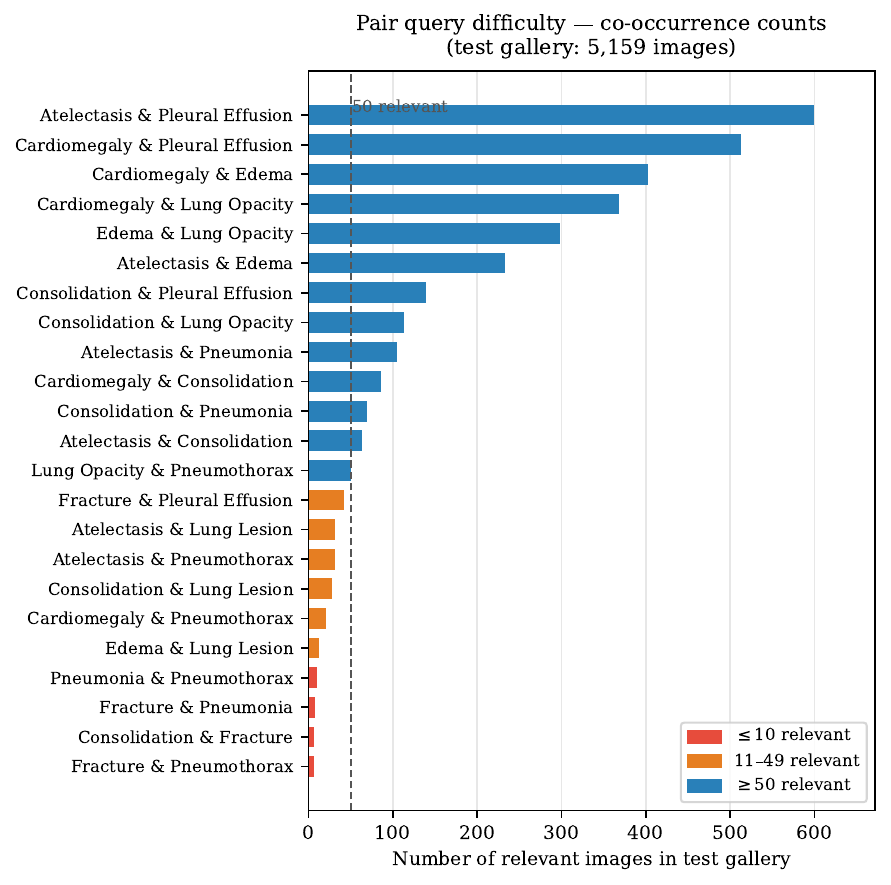}
\caption{Preview of relevant images per some pair query in the test gallery (5{,}159 images), sorted ascending. Half of all 45 queries have fewer than 50 relevant images, making conjunctive retrieval substantially harder than single-label queries.}
\label{fig:pair_difficulty}
\end{figure}

The largest gains appear on compositional retrieval. For conjunctive queries (\emph{A and B}), our model substantially outperforms all baselines, improving P@5 from 11.1 to 19.6 and R@5 from 33.3 to 51.1 relative to CXR-CLIP. Although absolute performance remains lower than in the single-label setting, this regime is considerably more challenging because many pathology pairs have only a small number of relevant images in the gallery (Fig.~\ref{fig:pair_difficulty} and Fig.~\ref{fig:cooccurrence}).

Negation retrieval is where the proposed objective provides the greatest benefit. Our model more than doubles CXR-CLIP's P@5 (42.0 vs. 20.0) and increases R@5 from 54.4 to 90.0. It also achieves the lowest HNRR@5 (1.6), indicating substantially fewer contradictory retrievals. These results suggest that explicitly modeling label conflicts and negations is critical for retrieving images that satisfy exclusion constraints.

Precision@5 comparison between models is shown in Fig ~\ref{fig:parallel_coords}.
Qualitative examples of composite query retrieval are shown in Fig.~\ref{fig:comparison_collage}, where our model retrieves images satisfying all query constraints, while CXR-CLIP fails across both conjunctive and negation queries. 

%--------------------------------------------------------------
\subsection{Negation Robustness}\label{subsec:negate}
%--------------------------------------------------------------
 
\begin{table}[t]
\centering
\caption{Negation robustness. \textbf{(a)} Performance under linguistic
variation, where the training template ``\textit{A} and no \textit{B}'' is
replaced at evaluation by ``an image with \textit{A} but without \textit{B}''.
\textbf{(b)} Hard Negative Retrieval Rate (lower is better): the fraction of
top-$k$ results that directly contradict the query.}
\label{tab:neg_robustness}
\begin{adjustbox}{width=\linewidth}
\begin{tabular}{lccccc}
\toprule
\multicolumn{6}{l}{\textit{(a) Paraphrased negation queries}} \\
\cmidrule(lr){1-6}
Model & P@1 & P@3 & P@5 & R@3 & R@5 \\
\midrule
\texttt{clip}       &  8.9 & 13.0 & 10.7 & 31.1 & 37.8 \\
\texttt{biomedclip} & 20.0 & 20.0 & 19.6 & 41.1 & 48.9 \\
\texttt{cxr-clip}   & 32.2 & 27.8 & 25.8 & 42.2 & 53.3 \\
\rowcolor{oursblue}
\texttt{ours}       & \textbf{43.3} & \textbf{43.3} & \textbf{41.6} & \textbf{73.3} & \textbf{85.6} \\
\bottomrule
\end{tabular}
\end{adjustbox}
 
\vspace{0.6em}
 
\begin{adjustbox}{width=\linewidth}
\begin{tabular}{lccc}
\toprule
\multicolumn{4}{l}{\textit{(b) Hard Negative Retrieval Rate (\%, $\downarrow$)}} \\
\cmidrule(lr){1-4}
Model & HNRR@1 & HNRR@3 & HNRR@5 \\
\midrule
\texttt{clip}       & 4.4 & 2.6 & 2.2 \\
\texttt{biomedclip} & 2.2 & 3.3 & 4.9 \\
\texttt{cxr-clip}   & 3.3 & 2.6 & 2.7 \\
\rowcolor{oursblue}
\texttt{ours}       & \textbf{1.1} & \textbf{1.1} & \textbf{1.6} \\
\bottomrule
\end{tabular}
\end{adjustbox}
\end{table}
 
The negation win is not an artifact of matching the training template. Under paraphrase (Tab.~\ref{tab:neg_robustness}a), our model retains 41.6 P@5 and 85.6 R@5---a marginal drop from the templated 42.0/90.0---while still leading CXR-CLIP by 15.8 points of P@5. Across both phrasings, and reinforced by the consistently low hard-negative rate (Tab.~\ref{tab:neg_robustness}b, HNRR@$k$ of 1.1--1.6 vs.\ 2.2--4.9 for the baselines), the model behaves as if it has learned a negation \emph{concept} rather than memorizing a surface string.

%--------------------------------------------------------------
\subsection{Ablations}\label{subsec:ablation}
%--------------------------------------------------------------
 
\begin{table}[t]
\centering
\caption{Ablations, all P@5 except the final HNRR@5 column ($\downarrow$). Each
block varies one design choice of \emph{our} pipeline; row names denote pipeline
variants, not the external baselines of Tab.~\ref{tab:main_results}. ``ours''
(repeated as the anchor) is the full method: CXR-CLIP init, 50-25-25 caption mix,
CLIP+Attract+HNM loss.}
\label{tab:ablation}
\begin{adjustbox}{width=\linewidth}
\begin{tabular}{lcccc}
\toprule
Variant & Single & Pair & Negation & HNRR@5\,$\downarrow$ \\
\midrule
\multicolumn{5}{l}{\textit{Loss formulation}} \\
\cmidrule(lr){1-5}
CLIP loss (diag.)        & 42.0 & 12.4 & 35.5 & 3.2 \\
\quad+ Attract           & \textbf{48.0} & 17.7 & 37.1 & 3.7 \\
\rowcolor{oursblue}
\quad+ HNM (\textbf{ours}) & \textbf{48.0} & \textbf{19.6} & \textbf{42.0} & \textbf{1.6} \\
\midrule
\multicolumn{5}{l}{\textit{Training caption mix (single--pair--negation)}} \\
\cmidrule(lr){1-5}
single-only & \textbf{58.0} & 18.7 & 34.0 & 3.2 \\
pair-only   & 38.0 & 15.1 & 20.9 &  3.7 \\
negation-only & 42.0 & \textbf{20.9} & \textbf{46.7} & \textbf{1.6} \\
\rowcolor{oursblue}
Mixed 2:1:1 ratio & 48.0 & 19.6 & 42.0 & \textbf{1.6} \\
\midrule
\multicolumn{5}{l}{\textit{Backbone initialization}} \\
\cmidrule(lr){1-5}
Open CLIP \cite{ilharco_gabriel_2021_5143773}        & \textbf{50.0} & 14.7 & 37.1 & 2.4 \\
\rowcolor{oursblue}
CXR-CLIP \cite{you2023cxrclip} & 48.0 & \textbf{19.6} & \textbf{42.0} & \textbf{1.6} \\
\bottomrule
\end{tabular}
\end{adjustbox}
\end{table}

 \vspace{-0.6em}
\paragraph{Loss formulation.}
Each loss component contributes where intended (table \ref{tab:ablation}). Replacing the diagonal-only CLIP
loss with label-aware attraction lifts pair P@5 from 12.4 to 17.7 and negation
from 35.5 to 37.1, but slightly worsens HNRR (3.2 to 3.7): attraction pulls
compatible pairs together without pushing contradictions apart. Adding hard
negative mining is what corrects this---HNRR drops sharply from 3.7 to 1.6 and
negation P@5 jumps from 37.1 to 42.0, at no cost to single-label
precision---confirming that the repulsion term does exactly the job it was
designed for.

\vspace{-0.6em}

\paragraph{Training caption mix.}
The 50-25-25 schedule is the best generalist \ref{tab:ablation}. Specialist mixes win their own
regime but collapse elsewhere: single-only peaks at 58.0 on single-label yet
falls to 18.7 on pairs and 34.0 on negation, while negation-only beats our
negation score but loses 6 points of single-label precision. No single
specialist matches our balance across all three regimes---training on every
query type is what buys robustness to every query type.

 \vspace{-0.6em}
\paragraph{Backbone initialization.}
In-domain initialization matters specifically for the hard queries. Fine-tuning from vanilla CLIP is competitive \ref{tab:ablation} on single-label (50.0, even edging our 48.0) but trails on pair (14.7 vs.\ 19.6), negation (37.1 vs.\ 42.0), and HNRR (2.4 vs 1.6). The compositional and negation gains ride on CXR-CLIP's in-domain visual features; our objective amplifies them rather than synthesizing them from scratch.

\vspace{-5pt}
\paragraph{Negation retrieval under alternative relevance definitions.}
When negation relevance is defined more strictly—requiring an explicitly confirmed negative label rather than treating unmentioned findings as absent—performance decreases substantially for all methods (Table~\ref{tab:negation_strict}). This reflects the scarcity of confirmed negative annotations in chest X-ray reports. Our model maintains a consistent advantage over CXR-CLIP, improving P@1 from 5.1 to 12.6 and P@5 from 5.3 to 8.6 while preserving a lower HNRR@5 (1.6 vs.\ 2.7), Which suggest that the gains observed in the main evaluation are not solely driven by treating unmentioned findings as negatives, but also extend to a more difficult setting.

\begin{table}
\centering
\caption{ Negation retrieval under alternative relevance definitions.
The main setting treats unmentioned findings as satisfying negation;
the strict setting requires a confirmed negative label.}
\label{tab:negation_strict}
\begin{adjustbox}{width=\linewidth}
\begin{tabular}{lcccc}
\toprule
Setting & Model & P@1 & P@5 & HNRR@5 \\
\midrule
Main & CXR-CLIP & 21.1  & 20.0 & 2.7 \\
\rowcolor{oursblue}
Main & Ours & \textbf{48.9} & \textbf{42.0}  & \textbf{1.6} \\
Strict confirmed negative & CXR-CLIP & 5.1 & 5.3 & 2.7 \\
\rowcolor{oursblue}
Strict confirmed negative & Ours & \textbf{12.6} & \textbf{8.6} & \textbf{1.6} \\
\bottomrule
\end{tabular}
\end{adjustbox}
\vspace{-5pt}

\end{table}

%Several other limitations remain for future work. First, our evaluation relies on CheXpert NLP labels; establishing a true clinical ceiling will require board-certified radiologist annotations. Second, because our models are trained and tested solely on MIMIC-CXR, out-of-distribution validation on external datasets is necessary to ensure cross-domain robustness against variations in scanner protocols and patient demographics.|% Finally, incorporating ranking-aware metrics (e.g., mAP, NDCG) alongside our threshold-based precision metrics will provide a more comprehensive evaluation of real-world search interface readiness.

\section{Conclusion}
In this work, we addressed the gap between traditional report-to-image matching and the demands of precision clinical retrieval. %\textsc{CXR-Retrieve} reveal this critical performance gap of existing models, while our label-aware contrastive objective demonstrate that explicitly modeling clinical assertions significantly improves performance on complex multi-pathology and negation queries. 
By introducing the \textsc{CXR-Retrieve} benchmark, we expose the severe limitations of current biomedical VLPs when faced with conjunctive and negation-based constraints. 
 
To bridge this gap, we proposed a label-aware contrastive fine-tuning method that aligns embeddings based on explicit clinical assertions rather than broad semantic similarity. Our approach demonstrates that actively repelling contradictory image-text pairs while attracting compatible clinical profiles yields substantial improvements in retrieval precision, as well as in our proposed HNRR metric. Ultimately, advancing medical image retrieval requires evaluating not just what visual findings a model can recognize, but how accurately it grasps the clinical logic binding them together.

Future work may explore other gaps such as less structured query text such as abbreviations (e.g., ``PTX'' for pneumothorax) or regional terminology, we deliberately excluded from this benchmark, to isolate the effects of severity performance gap caused by compositional and negation logic. Other aspects like efficincy of VLM \cite{wave}, image resolution \cite{cares} and spatial awareness\cite{awares}.
% A key limitation of our current framework is its reliance on formalized, structured query text. Real-world clinical queries are rarely this clean; physicians might use abbreviations (e.g., ``PTX'' for pneumothorax), regional terminology, and unstructured, messy descriptive constraints. While evaluating these ``wild'' text formulations is crucial for real-world deployment, we deliberately excluded them from this iteration of the benchmark. As our baselines demonstrate, even in a highly controlled and sanitized text setting, off-the-shelf and dedicated in-domain models still severely struggle with fundamental compositional and negation logic. By isolating these logical failure modes from natural language variation, our benchmark provides a clearer measure of a model's foundational constraint-satisfaction capabilities.

{
    \small
    \bibliographystyle{ieeenat_fullname}
    \bibliography{main}

\begin{thebibliography}{34}
\providecommand{\natexlab}[1]{#1}
\providecommand{\url}[1]{\texttt{#1}}
\expandafter\ifx\csname urlstyle\endcsname\relax
  \providecommand{\doi}[1]{doi: #1}\else
  \providecommand{\doi}{doi: \begingroup \urlstyle{rm}\Url}\fi

\bibitem[Alhamoud et~al.(2025)Alhamoud, Alshammari, Tian, Li, Torr, Kim, and Ghassemi]{alhamoud2025visionlanguagemodelsunderstandnegation}
Kumail Alhamoud, Shaden Alshammari, Yonglong Tian, Guohao Li, Philip Torr, Yoon Kim, and Marzyeh Ghassemi.
\newblock Vision-language models do not understand negation, 2025.

\bibitem[Alsentzer et~al.(2019)Alsentzer, Murphy, Boag, Weng, Jin, Naumann, and McDermott]{alsentzer2019publiclyavailableclinicalbert}
Emily Alsentzer, John~R. Murphy, Willie Boag, Wei-Hung Weng, Di Jin, Tristan Naumann, and Matthew B.~A. McDermott.
\newblock Publicly available clinical bert embeddings, 2019.

\bibitem[Cheng et~al.(2024)Cheng, Lange-Hegermann, H\"{o}vener, and Schreiweis]{Cheng2024}
Ka~Yung Cheng, Markus Lange-Hegermann, Jan-Bernd H\"{o}vener, and Bj\"{o}rn Schreiweis.
\newblock Instance-level medical image classification for text-based retrieval in a medical data integration center.
\newblock \emph{Computational and Structural Biotechnology Journal}, 24:\penalty0 434–450, 2024.

\bibitem[Cherti et~al.(2023)Cherti, Beaumont, Wightman, Wortsman, Ilharco, Gordon, Schuhmann, Schmidt, and Jitsev]{ilharco_gabriel_2021_5143773}
Mehdi Cherti, Romain Beaumont, Ross Wightman, Mitchell Wortsman, Gabriel Ilharco, Cade Gordon, Christoph Schuhmann, Ludwig Schmidt, and Jenia Jitsev.
\newblock Reproducible scaling laws for contrastive language-image learning.
\newblock In \emph{Proceedings of the IEEE/CVF Conference on Computer Vision and Pattern Recognition}, pages 2818--2829, 2023.

\bibitem[Endo et~al.(2021)Endo, Krishnan, Krishna, Ng, and Rajpurkar]{endo2021retrieval}
Mark Endo, Rayan Krishnan, Viswesh Krishna, Andrew~Y. Ng, and Pranav Rajpurkar.
\newblock Retrieval-based chest x-ray report generation using a pre-trained contrastive language-image model.
\newblock In \emph{Proceedings of Machine Learning for Health (ML4H)}, pages 209--219. PMLR, 2021.

\bibitem[Haq et~al.(2021)Haq, Moradi, and Wang]{HAQ2021101847}
Nandinee~Fariah Haq, Mehdi Moradi, and Z.~Jane Wang.
\newblock A deep community based approach for large scale content based x-ray image retrieval.
\newblock \emph{Medical Image Analysis}, 68:\penalty0 101847, 2021.

\bibitem[Hu et~al.(2021)Hu, Shen, Wallis, Allen-Zhu, Li, Wang, Wang, and Chen]{hu2021loralowrankadaptationlarge}
Edward~J. Hu, Yelong Shen, Phillip Wallis, Zeyuan Allen-Zhu, Yuanzhi Li, Shean Wang, Lu Wang, and Weizhu Chen.
\newblock Lora: Low-rank adaptation of large language models, 2021.

\bibitem[Huang et~al.(2021)Huang, Shen, Lungren, and Yeung]{huang2021gloria}
Shih-Cheng Huang, Liyue Shen, Matthew~P Lungren, and Serena Yeung.
\newblock Gloria: A multimodal global-local representation learning framework for label-efficient medical image recognition.
\newblock In \emph{Proceedings of the IEEE/CVF International Conference on Computer Vision}, pages 3942--3951, 2021.

\bibitem[Iglesias et~al.(2026)Iglesias, Talavera, and Troya]{IGLESIAS2026109540}
Guillermo Iglesias, Edgar Talavera, and Jesús Troya.
\newblock Chest x-ray deep learning comparative diagnosis using visual and semantic similarity with variational autoencoders.
\newblock \emph{Biomedical Signal Processing and Control}, 116:\penalty0 109540, 2026.

\bibitem[Irvin et~al.(2019)Irvin, Rajpurkar, Ko, Yu, Ciurea-Ilcus, Chute, Marklund, Haghgoo, Ball, Shpanskaya, et~al.]{irvin2019chexpert}
Jeremy Irvin, Pranav Rajpurkar, Michael Ko, Yifan Yu, Silviana Ciurea-Ilcus, Chris Chute, Henrik Marklund, Behzad Haghgoo, Robyn Ball, Katie Shpanskaya, et~al.
\newblock Chexpert: A large chest radiograph dataset with uncertainty labels and expert comparison.
\newblock In \emph{Thirty-Third AAAI Conference on Artificial Intelligence}, 2019.

\bibitem[Jeong et~al.(2023)Jeong, Tian, Li, Hartung, Behzadi, Calle, Osayande, Pohlen, Adithan, and Rajpurkar]{jeong2023multimodalimagetextmatchingimproves}
Jaehwan Jeong, Katherine Tian, Andrew Li, Sina Hartung, Fardad Behzadi, Juan Calle, David Osayande, Michael Pohlen, Subathra Adithan, and Pranav Rajpurkar.
\newblock Multimodal image-text matching improves retrieval-based chest x-ray report generation, 2023.

\bibitem[Johnson et~al.(2024)Johnson, Pollard, Mark, Berkowitz, and Horng]{PhysioNet-mimic-cxr-2.1.0}
Alistair Johnson, Tom Pollard, Roger Mark, Seth Berkowitz, and Steven Horng.
\newblock {MIMIC-CXR Database}.
\newblock \emph{{PhysioNet}}, 2024.
\newblock Version 2.1.0.

\bibitem[Johnson et~al.(2019)Johnson, Pollard, Greenbaum, Lungren, ying Deng, Peng, Lu, Mark, Berkowitz, and Horng]{johnson2019mimicjpg}
Alistair E.~W. Johnson, Tom~J. Pollard, Nathaniel~R. Greenbaum, Matthew~P. Lungren, Chih ying Deng, Yifan Peng, Zhiyong Lu, Roger~G. Mark, Seth~J. Berkowitz, and Steven Horng.
\newblock Mimic-cxr-jpg, a large publicly available database of labeled chest radiographs, 2019.

\bibitem[Kimhi et~al.(2024)Kimhi, Kimhi, Zheltonozhskii, Litany, and Baskin]{kimhi2024semi}
Moshe Kimhi, Shai Kimhi, Evgenii Zheltonozhskii, Or Litany, and Chaim Baskin.
\newblock Semi-supervised semantic segmentation via marginal contextual information, 2024.

\bibitem[Kimhi et~al.(2025)Kimhi, Koifman, Rivlin, Schwartz, and Baskin]{wave}
Moshe Kimhi, Erez Koifman, Ehud Rivlin, Eli Schwartz, and Chaim Baskin.
\newblock Waveclip: Wavelet tokenization for adaptive-resolution clip, 2025.

\bibitem[Kimhi et~al.(2026)Kimhi, Shabtay, Giryes, Baskin, and Schwartz]{cares}
Moshe Kimhi, Nimrod Shabtay, Raja Giryes, Chaim Baskin, and Eli Schwartz.
\newblock {CARES}: Context-aware resolution selector for {VLM}s.
\newblock In \emph{Proceedings of the 64th Annual Meeting of the {A}ssociation for {C}omputational {L}inguistics (Volume 1: Long Papers)}, pages 2243--2256, San Diego, California, United States, 2026. Association for Computational Linguistics.

\bibitem[Kumar et~al.(2013)Kumar, Kim, Cai, Fulham, and Feng]{Kumar2013}
Ashnil Kumar, Jinman Kim, Weidong Cai, Michael Fulham, and Dagan Feng.
\newblock Content-based medical image retrieval: A survey of applications to multidimensional and multimodality data.
\newblock \emph{Journal of Digital Imaging}, 26\penalty0 (6):\penalty0 1025–1039, 2013.

\bibitem[Litjens et~al.(2017)Litjens, Kooi, Bejnordi, Setio, Ciompi, Ghafoorian, van~der Laak, van Ginneken, and Sánchez]{Litjens_2017}
Geert Litjens, Thijs Kooi, Babak~Ehteshami Bejnordi, Arnaud Arindra~Adiyoso Setio, Francesco Ciompi, Mohsen Ghafoorian, Jeroen~A.W.M. van~der Laak, Bram van Ginneken, and Clara~I. Sánchez.
\newblock A survey on deep learning in medical image analysis.
\newblock \emph{Medical Image Analysis}, 42:\penalty0 60–88, 2017.

\bibitem[Liu et~al.(2021)Liu, Lin, Cao, Hu, Wei, Zhang, Lin, and Guo]{liu2021swintransformerhierarchicalvision}
Ze Liu, Yutong Lin, Yue Cao, Han Hu, Yixuan Wei, Zheng Zhang, Stephen Lin, and Baining Guo.
\newblock Swin transformer: Hierarchical vision transformer using shifted windows, 2021.

\bibitem[Murphy et~al.(2015)Murphy, Herrick, Wang, Wang, Sack, Andriole, Wei, Reynolds, Plesniak, Rosen, Pieper, and Gollub]{Murphy2015-zc}
Shawn~N Murphy, Christopher Herrick, Yanbing Wang, Taowei~David Wang, Darren Sack, Katherine~P Andriole, Jesse Wei, Nathaniel Reynolds, Wendy Plesniak, Bruce~R Rosen, Steven Pieper, and Randy~L Gollub.
\newblock High throughput tools to access images from clinical archives for research.
\newblock \emph{J. Digit. Imaging}, 28\penalty0 (2):\penalty0 194--204, 2015.

\bibitem[Peng et~al.(2017)Peng, Wang, Lu, Bagheri, Summers, and Lu]{peng2017negbiohighperformancetoolnegation}
Yifan Peng, Xiaosong Wang, Le Lu, Mohammadhadi Bagheri, Ronald Summers, and Zhiyong Lu.
\newblock Negbio: a high-performance tool for negation and uncertainty detection in radiology reports, 2017.

\bibitem[Qayyum et~al.(2017)Qayyum, Anwar, Awais, and Majid]{QAYYUM20178}
Adnan Qayyum, Syed~Muhammad Anwar, Muhammad Awais, and Muhammad Majid.
\newblock Medical image retrieval using deep convolutional neural network.
\newblock \emph{Neurocomputing}, 266:\penalty0 8–20, 2017.

\bibitem[Radford et~al.(2021)Radford, Kim, Hallacy, Ramesh, Goh, Agarwal, Sastry, Askell, Mishkin, Clark, et~al.]{radford2021learning}
Alec Radford, Jong~Wook Kim, Chris Hallacy, Aditya Ramesh, Gabriel Goh, Sandhini Agarwal, Girish Sastry, Amanda Askell, Pamela Mishkin, Jack Clark, et~al.
\newblock Learning transferable visual models from natural language supervision.
\newblock In \emph{International conference on machine learning}, pages 8748--8763. PMLR, 2021.

\bibitem[Ray et~al.(2023)Ray, Radenovic, Dubey, Plummer, Krishna, and Saenko]{ray2023cola}
Arijit Ray, Filip Radenovic, Abhimanyu Dubey, Bryan~A Plummer, Ranjay Krishna, and Kate Saenko.
\newblock Cola: A benchmark for compositional text-to-image retrieval.
\newblock \emph{arXiv preprint arXiv:2305.02882}, 2023.

\bibitem[Shabtay et~al.(2026)Shabtay, Kimhi, Spector, Haray, Rivlin, Baskin, Giryes, and Schwartz]{awares}
Nimrod Shabtay, Moshe Kimhi, Artem Spector, Sivan Haray, Ehud Rivlin, Chaim Baskin, Raja Giryes, and Eli Schwartz.
\newblock Look where it matters: High-resolution crops retrieval for efficient vlms.
\newblock \emph{arXiv preprint arXiv:2603.16932}, 2026.

\bibitem[Sotomayor et~al.(2021)Sotomayor, Mendoza, Castañeda, Farías, Molina, Pereira, H\"{a}rtel, Solar, and Araya]{Sotomayor2021}
Camilo~G. Sotomayor, Marcelo Mendoza, Víctor Castañeda, Humberto Farías, Gabriel Molina, Gonzalo Pereira, Steffen H\"{a}rtel, Mauricio Solar, and Mauricio Araya.
\newblock Content-based medical image retrieval and intelligent interactive visual browser for medical education, research and care.
\newblock \emph{Diagnostics}, 11\penalty0 (8):\penalty0 1470, 2021.

\bibitem[Thrush et~al.(2022)Thrush, Jiang, Bartolo, Singh, Williams, Kiela, and Ross]{thrush2022winoground}
Tristan Thrush, Ryan Jiang, Max Bartolo, Amanpreet Singh, Adina Williams, Douwe Kiela, and Candace Ross.
\newblock Winoground: Probing vision and language models for visio-linguistic compositionality.
\newblock In \emph{Proceedings of the 2022 Conference on Empirical Methods in Natural Language Processing (EMNLP)}, 2022.

\bibitem[Wang et~al.(2017)Wang, Peng, Lu, Lu, Bagheri, and Summers]{Wang_2017}
Xiaosong Wang, Yifan Peng, Le Lu, Zhiyong Lu, Mohammadhadi Bagheri, and Ronald~M. Summers.
\newblock Chestx-ray8: Hospital-scale chest x-ray database and benchmarks on weakly-supervised classification and localization of common thorax diseases.
\newblock In \emph{2017 IEEE Conference on Computer Vision and Pattern Recognition (CVPR)}, page 3462–3471. IEEE, 2017.

\bibitem[Wang et~al.(2022)Wang, Wu, Agarwal, and Sun]{wang2022medclipcontrastivelearningunpaired}
Zifeng Wang, Zhenbang Wu, Dinesh Agarwal, and Jimeng Sun.
\newblock Medclip: Contrastive learning from unpaired medical images and text, 2022.

\bibitem[You et~al.(2023)You, Gu, Ham, Park, Kim, Hong, Baek, and Roh]{you2023cxrclip}
Kihyun You, Jawook Gu, Jiyeon Ham, Beomhee Park, Jiho Kim, Eun~K. Hong, Woonhyuk Baek, and Byungseok Roh.
\newblock \emph{CXR-CLIP: Toward Large Scale Chest X-ray Language-Image Pre-training}, page 101–111.
\newblock Springer Nature Switzerland, 2023.

\bibitem[Yuksekgonul et~al.(2023)Yuksekgonul, Bianchi, Kalluri, Jurafsky, and Zou]{yuksekgonul2023visionlanguagemodelsbehavelike}
Mert Yuksekgonul, Federico Bianchi, Pratyusha Kalluri, Dan Jurafsky, and James Zou.
\newblock When and why vision-language models behave like bags-of-words, and what to do about it?, 2023.

\bibitem[Zhang et~al.(2023)Zhang, Xu, Usuyama, et~al.]{biomedclip}
Sheng Zhang, Yanbo Xu, Naoto Usuyama, et~al.
\newblock Biomedclip: a multimodal biomedical foundation model pretrained from fifteen million scientific image-text pairs.
\newblock \emph{arXiv preprint arXiv:2303.00915}, 2023.

\bibitem[Zhang et~al.(2025)Zhang, Zhao, Wu, Zhou, Zhang, Wang, and Xie]{zhang2025radirscalableframeworkmultigrained}
Tengfei Zhang, Ziheng Zhao, Chaoyi Wu, Xiao Zhou, Ya Zhang, Yanfeng Wang, and Weidi Xie.
\newblock Radir: A scalable framework for multi-grained medical image retrieval via radiology report mining, 2025.

\bibitem[Zhang et~al.(2022)Zhang, Jiang, Miura, Manning, and Langlotz]{zhang2022contrastivelearningmedicalvisual}
Yuhao Zhang, Hang Jiang, Yasuhide Miura, Christopher~D. Manning, and Curtis~P. Langlotz.
\newblock Contrastive learning of medical visual representations from paired images and text, 2022.

\end{thebibliography}
}

% \newpage
% \appendix

% \begin{table}[ht]
% \centering
% \caption{Per-image label count breakdown: fraction of images with zero, one or more than one label of each type.}
% \label{tab:per_row_breakdown}
% \begin{tabular}{lrrr}
% \hline
% \textbf{Label} & \textbf{=0} & \textbf{=1} & $\mathbf{>1}$ \\
% \hline
% Positive ($1$)   & 45.3\%  & 24.2\% & 30.5\% \\
% Negative ($0$)   & 64.3\% & 19.8\% & 15.8\% \\
% Uncertain ($-1$) & 80.3\% & 14.4\% & 5.3\% \\
% Missing (NaN)    & 0.0\%  & 0.0\%  & 100.0\% \\
% \hline
% \end{tabular}
% \end{table}

\end{document}